%% file: root.tex
\documentclass{article}
\usepackage{hyperref}
\usepackage[final,nohyperref]{corl_2018}
\usepackage{amsmath}
\usepackage{amsfonts}
\input{variables.tex}
\usepackage{enumitem}
\setitemize{noitemsep,topsep=0pt,parsep=0pt,partopsep=0pt}

\title{Modular meta-learning}

\author{
  Ferran Alet, Tom\'{a}s Lozano-P\'{e}rez, Leslie P. Kaelbling\\
  MIT Computer Science and Artificial Intelligence Laboratory\\
  \texttt{\{alet,tlp,lpk\}@mit.edu} \\
}

\newcommand{\OUR}{{\sc BounceGrad}}
\newcommand{\algname}{{\sc BounceGrad}}
\newcommand{\algMAML}{{\sc MOMA}}

\newcommand{\hypoths}{({\cal C}, F, \Theta)}
\newcommand{\MAML}{{\sc MAML}}
\newcommand{\dtrain}{D^\text{\it train}}
\newcommand{\dval}{D^\text{\it test}}

\begin{document}
\maketitle

\vspace{-\baselineskip}
\begin{abstract}
Many prediction problems, such as those that arise in the context of robotics, have a simplifying underlying structure that, if known, could accelerate learning. In this paper, we present a strategy for learning a set of neural network modules that can be combined in different ways. We train different modular structures on a set of related tasks and generalize to new tasks by composing the learned modules in new ways. By reusing modules to generalize we achieve \textit{combinatorial generalization}, akin to the "infinite use of finite means" displayed in language. Finally, we show this improves performance in two robotics-related problems.
\end{abstract}
\vspace{-\baselineskip}

\section{Introduction}

In many situations, such as robot learning, training experience is very expensive.  One strategy for reducing the amount of training data needed for a new task is to learn some form of prior or bias using data from several related tasks.  The objective of this process is to extract information that will substantially reduce the training-data requirements for a new task.  
This problem is a form of transfer learning, sometimes also called meta-learning or ``learning to learn'' \citep{schmidhuber1987evolutionary,thrun2012learning}.

Previous approaches to meta-learning have focused on finding distributions over \citep{hierarchicalBayes} or initial values of~\citep{MAML,Reptile} parameters, based on a set of ``training tasks,'' that will enable a new ``test task'' to be learned with many fewer training examples.  Our objective is similar, but rather than focusing on transferring information about parameter values, we focus on finding a set of reusable modules that can form components of a solution to a new task, possibly with a small amount of tuning. By reusing our learned modules, we aim at \textit{combinatorial generalization}\cite{humboldt1999language,chomsky2014aspects,battaglia2018relational}; this is akin to the reuse of words to construct many possible sentences. We propose that this "infinite use of finite means" (Von Humboldt) can be a scalable approach towards transfer and generalization.

Modular approaches to learning have been very successful in structured tasks such as natural-language sentence interpretation \citep{andreas2016neural}, in which the input signal gives relatively direct information about a good structural decomposition of the problem.  We wish to address problems that may benefit from a modular decomposition but do not provide any task-level input from which the structure of a solution may be derived.  Nonetheless, 
we adopt a similar modular structure and parameter-adaptation method for learning reusable modules, but use a general-purpose simulated-annealing search strategy to find an appropriate structural decomposition for each new task.

We provide an algorithm, called \algname{}, which learns a set of modules and then combines them appropriately for a new task.  We demonstrate its effectiveness by comparing it to \MAML{}~\citep{MAML}, a popular meta-learning method, on a set of regression tasks that represent the types of prediction-learning problems that are faced by robotics systems, and show that we achieve better prediction performance from a few training examples, and can be much faster to train.
In addition, we show that this modular approach offers a strategy for explaining learned solutions to new tasks:  by observing the modules that are used in a new task, we can relate this task to previous tasks that use the same modules.  This approach also offers opportunities for verification and validation:  the modules discovered during meta-learning may be subjected to computationally expensive analytical or empirical validation techniques off-line;  they can then be recombined to address new tasks, generating solutions that can be validated more efficiently as compositions of previously understood modules.
\vspace{-.5\baselineskip}
\section{Related Work}

\begin{figure}
    \centering
    \includegraphics[width=.8\textwidth]{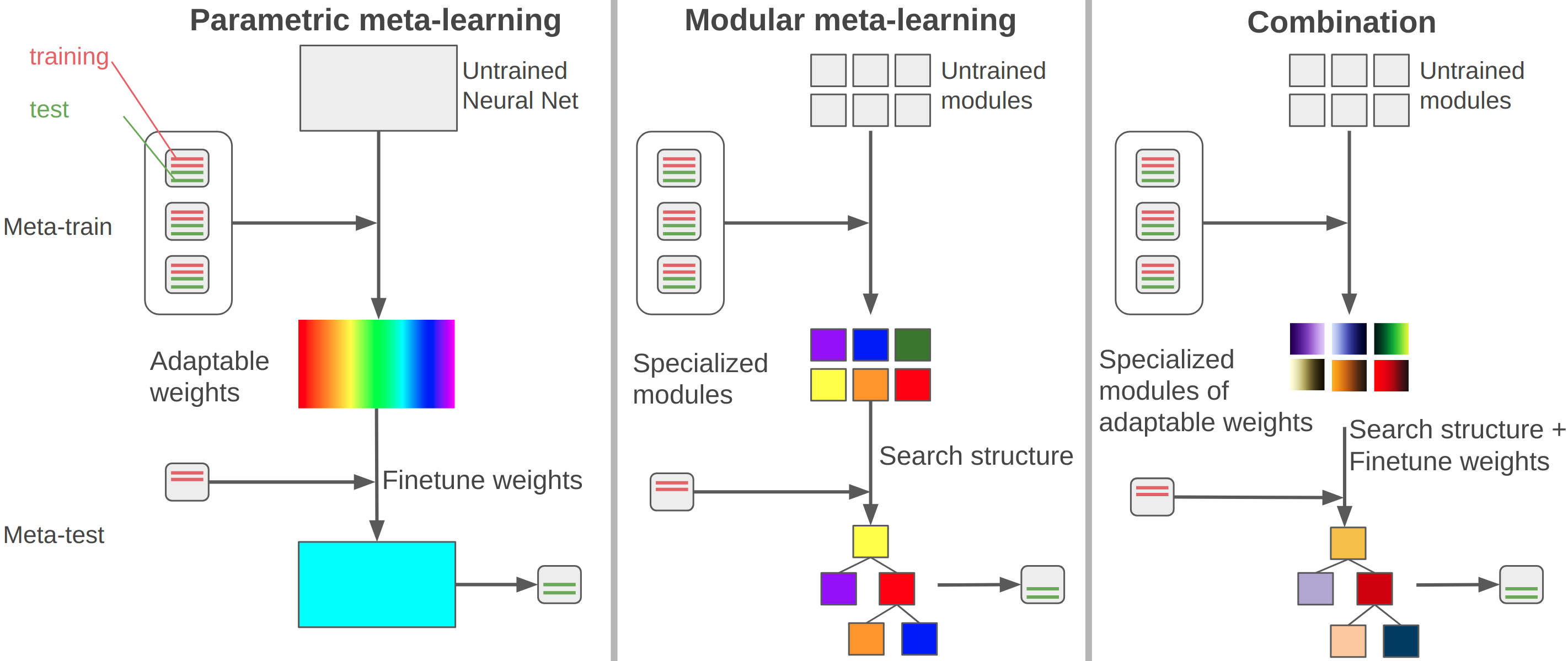}
    \caption{All methods train on a set of related tasks and obtain some flexible intermediate representation.  Parametric strategies such as \MAML{} (left) learn a representation that can be quickly adjusted to solve a new task.  Our modular meta-learning method (middle) learns a repertoire of modules that can be quickly recombined to solve a new task.  A combination of \MAML{} and modular meta-learning (right) learn initial weights for modules that can be combined and adapted for a new task.
    \label{fig:Leslie}}
    \vspace{-1.5\baselineskip}
\end{figure}

Our work draws primarily from two sources: multi-task meta-learning and modular networks.
Prominent examples of meta-learning in robotic domains are \MAML{}~\citep{MAML} and follow-up work~\citep{Reptile, yu2018one}. They perform ``meta-training'' on a set of related tasks with the goal of finding network weights that serve as a good {\em starting point} for a few steps of gradient descent in each task. 
Others \citep{duan2017one,duan2016rl,andrychowicz2016learning,bengio1990learning,edwards2016towards} perform different types of parametric changes in the network's computation conditioned on few examples.
 We adapt the same basic setting, but rather than finding good starting weights, we find a good set of modules for later structural combination; see figure \ref{fig:Leslie}. This is akin to the distinction in AI and cognitive science between parameter change vs. structural change~\citep{tenenbaum2011grow,ullman2018learning}.
 
Neural module networks~\citep{andreas2016neural} provide an elegant mechanism for training a set of individual modules that can be recombined to solve new problems, when the input has enough signal to guess an appropriate modular decomposition. ~\citet{johnson2017inferring} later showed that the structure controller could be trained with RL; others applied similar frameworks to get more interpretability~\citep{al2017contextual} or to generalize across robotic tasks with neural programs\citep{xu2017neural}. However, as far as we know, this framework has not been applied in problems where the input does not give enough information about an appropriate structure.

Structured networks have been used for meta-learning in the reinforcement-learning setting. \citet{devin2017learning} use a fixed composition of two functions, one related to the robot and one to the task. 
\citet{frans2017meta} jointly optimize a master policy and a set of sub-policies (options) that can be used to solve new problems; this method can be seen as having a single fixed scheme of combination via the master policy;  it is in contrast to our ability to handle a variety of computational compositions.  
{\sc PathNet}~\citep{fernando2017pathnet} is closely related to our work.  The architecture is layered, with several modules in each layer.  An evolutionary algorithm determines gates on the connections among the modules. After training on an initial task, the weights in the modules that contribute to the solution to that task are frozen, and then the architecture is trained on a second task. If the tasks are sufficiently related, the modules learned in the first task can be directly re-used to make learning more efficient in the second task.  
\citet{meyerson2017beyond} and later \citet{liang2018evolutionary} expanded these ideas to the multitask setting with two particular compositional schemes: soft layer ordering and routing in DAGs. We propose a general framework of which these are two important sub-cases. Moreover, we operate in the meta-learning setting where, with few points per task, it is very easy to prematurely optimize the structure and run into local optima, as shown in figure \ref{fig:comparative_advantage}. Therefore, we believe using simulated annealing rather than gradient descent\citep{meyerson2017beyond} or irreversible evolution\citep{liang2018evolutionary} may be a better fit for our setting.


\section{Modular meta-learning}
We explore the problem of modular meta-learning in the context of {\em supervised learning} problems, in which training and validation sets of input-output pairs are available.  Such problems arise in robotics, particularly in learning to predict a next state based on information about previous states and actions.  
We believe that techniques similar to ours can be applied to reinforcement-learning problems as well, but do not explore that in this paper.
We use the same meta-learning problem formulation as Finn et al.\citep{MAML} used to apply \MAML{} to supervised learning problems.  
We assume an underlying distribution $p({\cal T})$ over tasks:  a task is a joint distribution $P_{\cal T}(x, y)$ over $(x, y)$ pairs.
The learning problem is:  Given data drawn from $m$ meta-training tasks and a (small) amount of data drawn from a {\em meta-test} task,  where all tasks are drawn from $P({\cal T})$, find a hypothesis $h$ from some given set that incurs low expected loss on unseen data drawn from the meta-test task.  In this formulation, each task is characterized by two data sets, $\dtrain$ and $\dval$, each of which consists of a set of input-output pairs $(x,y)$.  We assume a differentiable loss function ${\cal L}(\hat{y}, y)$ on true vs predicted outputs for all tasks. 

\subsection{Structured hypotheses}

\label{sec:compositions}

We begin by defining a family of structured hypothesis classes.  Given the specification of a composition rule and a basis set of modules, $\hypoths{}$\add{may want to change this sentence or order in hypoths} represents a set of possible functional input-output mappings that will serve as the hypothesis space for the meta-test task. 
$F$ is a basis set of modules, which are functions $f_1, \ldots, f_k$; each function has a parametric form $y = f_i(x; \theta_i)$ where $\theta_i$ is a fixed-dimensional vector of parameters.  In this work, all the $f_i$ are neural networks, potentially with different architectures, and the parameters $\Theta = (\theta_1, \ldots, \theta_k)$ are the weights of the neural networks, which differ among the modules.
${\cal C}$ is a {\em compositional scheme} for forming complex functions from simpler ones, defined by an initial structure 
and a set of local modification operations on the structure.  Some examples include:
    \begin{itemize}[noitemsep,topsep=0pt,parsep=0pt,partopsep=0pt,leftmargin=*]
    \item Single module $h(x) = f_i(x)$, as fig. \ref{fig:comparative_advantage}.  The local modification is to change which module is used.  
    \item A fixed compositional structure, e.g., $h(x) = f_i(x) + f_j(x)$ or $h(x) = f_i(f_j(x))$.  The local modifications are to change which module is used for each of the component functions. We could generalize this to stacking many CNN/\textit{ResNet} layers \cite{he2016deep} for meta-learning in vision problems.
    \item A weighted ensemble, of the same basic form as an attention mechanism:
    \[h(x) = \sum_{l=1}^m \frac{e^{f_{i_l}(x)}}{\sum_{l'=1}^m e^{f_{i_{l'}}(x)}} g_{j_l}(x) \]
    where $i_1, \ldots, i_m$ and $j_1, \ldots, j_m$ are elements of the set $\{1, \ldots, k\}$, picking out which modules to use to play these roles in the network.  There are modules of two types:  the $f_i$ have a scalar output and the $g_i$ have an output that is the output dimension of the main regression problem.  The local modifications are to change which particular $f$ and $g$ modules are used for each role.  
    \item A general function-composition tree, where the local modifications include both changing which $f_i$ is used at each node, as well as adding or deleting nodes from the tree.
    \end{itemize}
Let ${\mathbb S}$ be the set of possible structures and $S \in {\mathbb S}$ be a particular structure, generated by ${\cal C}$, including a choice of which particular functions $f_i \in F$ are included in the structure.
To formulate a structured-hypothesis model, 
we must specify the number and parametric forms, but not parameter values, of basis functions, $F$, and compositional scheme ${\cal C}$.  This is analogous to specifying the architecture of a deep neural network.

Our approach has two phases:  an off-line {\em meta-learning} phase and an on-line {\em meta-test learning} phase.  In the meta-learning phase, we take training and validation data sets for tasks  $1, \ldots, k$ as input and generate a parametrization for each module, $\Theta = (\theta_1, \ldots,  \theta_k)$ as output; the objective is to construct modules that will work together as good building blocks for future tasks.
In the meta-test learning phase, we take a training data set for the meta-test task as input, as well as ${\mathbb S}$ and $\Theta$;  the output is a compositional form $S\in{\mathbb S}$ which includes a selection of modules $f_1 \ldots, f_{m_s}$ to be used in that form (a single element $f_j \in F$ may occur multiple times in $S$).  Since $\Theta$ is already specified, the choice of $S$ completely determines a mapping from inputs to outputs;  we will abuse notation slightly and write $S_\Theta$ to stand for the function from input to output generated by structure $S$ and parameters $\Theta$.
We may optionally include a meta-test tuning phase, which will adapt the parameter vectors; this is discussed in section~\ref{sec:moma}.

At learning time on the meta-test task, the space of possible structures ${\mathbb S}$ and parameters $\Theta$ are fixed, and the objective is to find and return the best structure in ${\mathbb S}$.
Define $e(D, S, \Theta)$ to be the loss of the hypothesis $S_\Theta$ on data set $D$:
$e(D, S, \Theta) = \sum_{\{(x, y) \in D\}} {\cal L}(S_\Theta(x), y)$.
Then our hypothesis is
\begin{equation}
S^*_\Theta = {\rm arg}\min_{S \in {\mathbb S}} e(\dtrain_\text{\it meta-test}, S, \Theta)
\label{eq:finalObjective}
\end{equation}
The hope is that, by choosing a limited but flexible and appropriate hypothesis space based on previous tasks, a good choice of $S^*_\Theta$ can be made based on a small amount of data in $\dtrain_\text{\it meta-test}$.

At meta-learning time, ${\mathbb S}$ is specified, and
the objective is to find parameter values $\Theta$ that constitute a set of modules that can be recombined to effectively solve each of the training tasks.  
We use validation sets for the meta-training tasks to avoid choosing $\Theta$ in a way that over-fits.
Our training objective is to find $\Theta$ that minimizes the average generalization performance of the hypotheses chosen by equation~\ref{eq:finalObjective} using parameter set $\Theta$: 
\begin{equation}
J(\Theta) = \sum_{j = 1}^m e(\dval_j, {\rm arg} \min_{S \in {\mathbb S}} e(\dtrain_j, S, \Theta), \Theta)\;\;.
\label{eq:metaObjective}
\end{equation}
\subsection{Learning algorithm}
The optimization problems specified by equations~\ref{eq:finalObjective} and~\ref{eq:metaObjective} are in general quite difficult, requiring a mixed continuous-discrete search in a space with many local optima.  
In this section, we describe the \algname{} algorithm, which performs local searches based on a combination of simulated annealing and gradient descent to find approximately optimal solutions to these problems.

\subsubsection{Meta-test learning phase}

In the meta-test learning phase, we have fixed the parameters $\Theta$ and only need to find an optimal structure $S \in {\mathbb S}$ according to the objective in equation~\ref{eq:finalObjective}.  We use simulated annealing~\citep{kirkpatrick1983optimization} to perform this search:  it is a local search strategy that uses stochasticity to avoid local optima and has asymptotic optimality guarantees.  We start with an initial structure, then randomly propose structural changes using local modification operators associated with the compositional scheme ${\mathbb S}$, accept the change if it decreases the error on the task and, with some probability, even if it does not.
\begin{algorithmic}
\Procedure{Online}{$\dtrain_{\it meta-test}$, ${\mathbb S}$, $\Theta$, $T_0$, $\Delta_T$, $T_{\it end}$}
\State $S = \text{random simple structure from ${\mathbb S}$}$
\For{$T = T_0;\; T = T - \Delta_T;\; T < T_{\it end}$}
        \State $S' = \text{\sc Propose}_{\mathbb S}(S)$
        \If{$\text{\sc Accept}(e(D, S', \Theta), e(D, S, \Theta), T)$} $S = S'$
        \EndIf
\EndFor        
\State \textbf{return} $S$
\EndProcedure
\Procedure{Accept}{$v'$, $v$, $T$}
    \State \textbf{return} $v' < v$ or {\rm rand}(0, 1) $<\exp\{(v - v')/T\}$
\EndProcedure
\end{algorithmic}
In order for simulated annealing to converge, the {\em temperature} parameter $T$ must be decreased over time.
The schedule we use decreases too quickly to satisfy theoretical convergence guarantees, yet is practically effective. 
Given the training set for the meta-test task, we run $\text{\sc Online}(\dtrain_{\it meta-test}, {\mathbb S}, \Theta)$ to obtain a hypothesis for that task.

\subsubsection{Meta-learning phase}


To perform the optimization in equation~\ref{eq:metaObjective}, we might use an algorithm that, in the outer loop, performs optimization over continuous parameters $\Theta$, where the evaluation of $\Theta$ consists of running procedure {\sc Online} on each of the training data sets, and evaluating the resulting structural hypotheses using the validation sets.  This strategy is ineffective because of the prevalence of bad local optima in the space, as illustrated in figure \ref{fig:comparative_advantage}.
As in clustering, we can smooth out some local optima by changing the objective function, although we will do so only during search, so our meta-test objective will remain the same.  We formulate a smoothed objective
\begin{equation}
\widehat{J}(\Theta, T) = \sum_{j = 1}^m \mathbb{E}_{S \sim {\rm MC}({\mathbb S}, v(s;\Theta), T)} e(\dval_j, S, \Theta)
\label{eq:smoothObjective}
\end{equation}
Here, ${\rm MC}({\mathbb S}, v, T)$ is the Markov chain induced by executing the simulated-annealing sampler in the structure space ${\mathbb S}$ using its proposal operator, with score function $v(s;\Theta)=e(\dtrain_j, s, \Theta)$ and {\em fixed} temperature $T$.  Rather than trying to find the $\Theta$ values that work best when we choose the best structure $S$, we will instead try to find $\Theta$ values that work best in expectation given the distribution of structures induced by the Markov chain. This space is smoother and less susceptible to local optima.  At the same time as we are optimizing $\Theta$ via stochastic gradient, we will cool the temperature of the Markov chain.  As $T$ approaches $0$, the objective $\widehat{J}$ becomes the same as our original objective $J$.


 Given particular structures $S_j$, then for each task $T_j$, we know the parametric form of a differentiable feed-forward function that has parameters drawn from $\Theta$, but possibly with parameter tying within and across the tasks due to re-use of the basis functions in $F$ and possibly with some parameters in $\Theta$ left unused.  We can adjust $\Theta$ using stochastic gradient descent, as defined in procedure {\sc Grad}.  It takes the structures and training data as input, as well as a step size $\eta$ and performs one step of standard stochastic gradient descent, or any alternative optimizer:
\begin{algorithmic}
\Procedure{Grad}{$\Theta$, $S_1, \ldots, S_m$, $\dval_1, \ldots, \dval_m$, $\eta$}
    \State $\Delta = 0$
    \For{$j = 1 \ldots m$}
        \State $(x, y) = {\rm rand\_elt}(\dval_j)$;\;\;
         $\Delta = \Delta + \nabla_\Theta L({S_j}_\Theta(x), y)$
    \EndFor
    \State $\Theta = \Theta - \eta \Delta$
\EndProcedure
\end{algorithmic}
However, we do not know the appropriate structure for each task, and so, according to the smoothed criterion in equation~\ref{eq:smoothObjective}, we sample structures using a stochastic process based on simulated annealing.  
We define a procedure {\sc Bounce} that takes a simulated annealing step on a structural hypothesis for each task, using the current $\Theta$ values, for a fixed temperature $T$:
\begin{algorithmic}
\Procedure{Bounce}{$S_1, \ldots, S_m$, $\dtrain_1, \ldots, \dtrain_m$,$T$, ${\mathbb S}$, $\Theta$}
    \For{$j = 1 \ldots m$}      
        \State $S'_j = {\sc Propose}_{\mathbb S}(S_j, \Theta)$
        \If{${\sc Accept}(e(\dtrain_j, S'_j, \Theta), e(\dtrain_j, S_j, \Theta), T)$}  $S_j = S'_j$
        \EndIf
    \EndFor
\EndProcedure
\end{algorithmic}
Finally, we combine these methods into an overall algorithm for optimizing equation~\ref{eq:finalObjective} via optimizing equation~\ref{eq:smoothObjective} and decaying $T$:
\begin{algorithmic}
\Procedure{BounceGrad}{${\mathbb S}$, $\dtrain_1, \ldots, \dtrain_m$,
$\dval_1, \ldots, \dval_m$,
$\eta$, $T_0, \Delta_T, T_{\it end}$}
\State $S_1, \ldots, S_m = \text{random simple structures from ${\mathbb S}$}$; \;\;
 $\Theta = \text{neural-network weight initialization}$
\For{$T = T_0;\; T = T - \Delta_T;\; T < T_{\it end}$}
    \State {\sc Bounce}($S_1, \ldots, S_m$, $\dtrain_1, \ldots, \dtrain_m$, $T$, ${\mathbb S}$, $\Theta$)
    \State {\sc Grad}($\Theta$, $S_1, \ldots, S_m$, $\dval_1, \ldots, \dval_m$, $\eta$)
\EndFor
\EndProcedure
\end{algorithmic}

\begin{figure}[t]
    \centering
    \subfloat[\label{fig:induced_distributions}]
    {\includegraphics[width=0.7\textwidth]{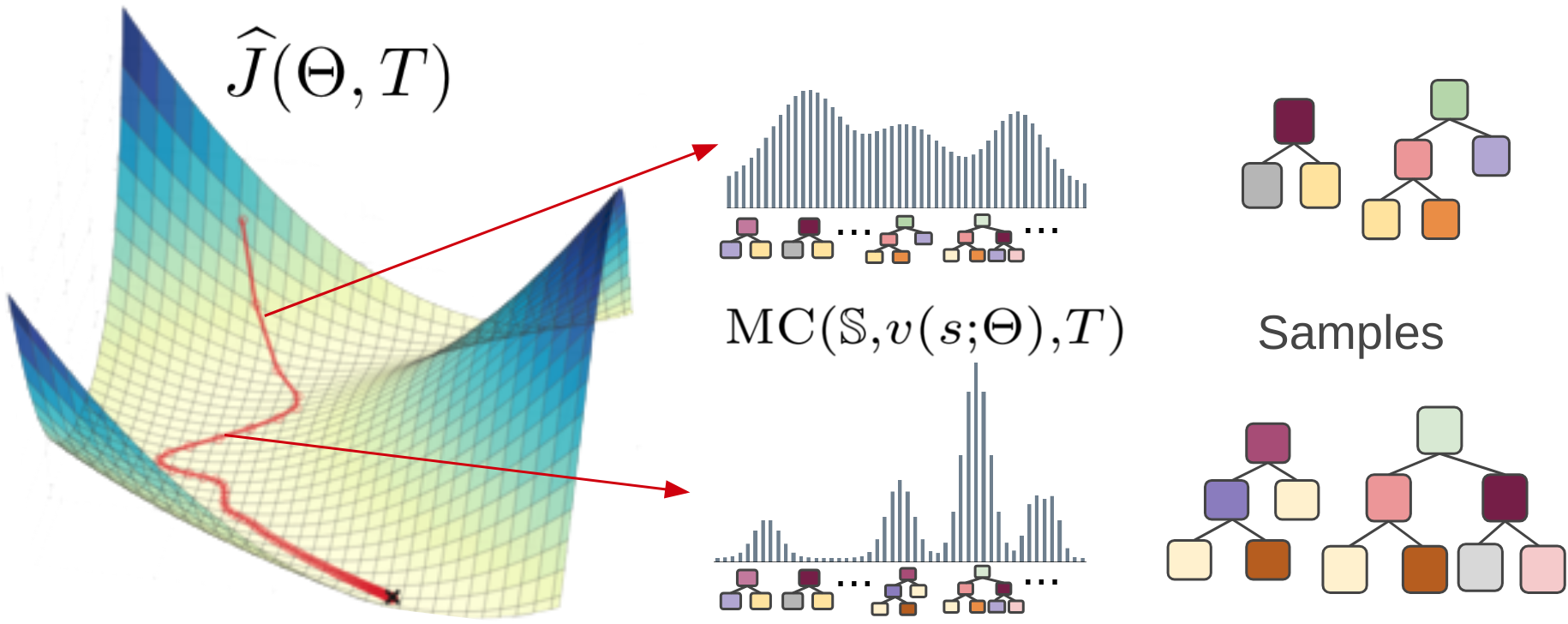}}\hfill
    \subfloat[ \label{fig:robotic_domains}]
    {\includegraphics[width=0.295\textwidth]{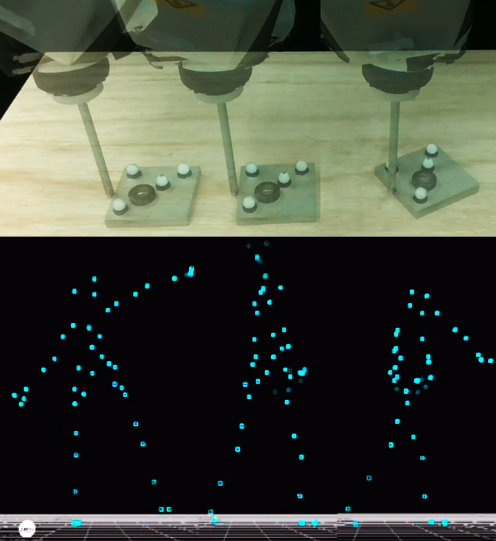}}\hfill
    \caption{(a) A schematic view of the optimization landscape over $\Theta$ and $T$.  At each point, there is a distribution over structures.  As temperature decreases, the variance of the distribution decreases and modules become more specialized. (b) benchmarked robotic domains: MIT push dataset~\citep{yu2016more}(top), action (throwing a ball) in the Berkeley MHAD dataset~\citep{ofli2013berkeley} (bottom).}
    \label{fig:induced}
\vspace{-1.5\baselineskip}
\end{figure}
We can think of the state of the optimization algorithm as consisting of both the parameters $\Theta$ and the temperature $T$;  these values induce a distribution on structures.  The optimization landscape is illustrated in figure~\ref{fig:induced_distributions}.  At high temperatures, the distribution over structures is diffuse and the modules will tend to be very generalized.  As the temperature decreases, modules can specialize to perform well at the roles they are being selected to play in the structures.


\subsection{Parameter tuning in online phase}
\label{sec:moma}
In the basic {\sc Online} method, we search for the best structure for the new task, without modifying parameters $\Theta$.  In fact, in many cases it may be useful to perform some additional parameter optimization as well.  One strategy would be to proceed as we have described before, running \algname{} on the training tasks to get $\Theta$, finding the best $S$ for the meta-test task using {\sc Online}, and then taking some gradient steps on $\Theta$, given $S$, to better optimize loss on $\dtrain$.  
However, we can do better by incorporating the objective of {\sc MAML} more deeply into both {\sc Online} and \algname{}, by redefining the inner error function used in the optimization criterion for $\Theta$:  instead of using $\Theta$ directly, we will evaluate the result of taking one (or a few) gradient steps away from $\Theta$, specialized to optimize $D$.  So, 
$e_\text{\sc maml}(D, S, \Theta) = \sum_{\{(x, y) \in D\}} L(S_{O(\Theta, D, S)}(x), y)$,
where the {\em optimized} parameters $O(\Theta, D, S)$ are obtained by a gradient update: 
$O(\Theta, D, S) = \Theta - \eta \nabla_\Theta e(D, S, \Theta)$.
Then, the meta-learning objective becomes
\begin{equation}
J_\text{\sc maml}(\Theta) = \sum_{j = 1}^m e(\dval_j, {\rm arg} \min_{S \in {\mathbb S}} e_\text{\sc maml}(\dtrain_j, S, \Theta), \Theta)
\label{eq:metaMamlObjective}
\end{equation}
We can therefore use $e_\text{\sc maml}$ in place of $e$ in the {\sc Grad} and {\sc Online} procedures, and perform a few additional gradient steps on $\Theta$ after obtaining the structure from {\sc Online}.  We will call this overall algorithm \algMAML{}, MOdular MAml.

\section{Experiments}
We compare four different learning approaches:  training a single network using the combined data from all tasks ({\sc Pooled}), training a single network using the {\sc MAML} criterion, training a modular network without weight adaptation in the online training (\algname{}), and training a modular network with {\sc MAML} adaptation in the online training (\algMAML{}).
To make the comparisons as fair as possible, for experiments on a given data set, all networks have the same shape: generally, a feedforward structure of 3--4 layers. We selected a set of layer sizes so that {\sc Pooled} and {\sc maml} had about 10 times as many parameters as the structured methods, to compensate for \algname{} and \algMAML{} having about 10 modules to combine. We also verified that none of the algorithms' performance was sensitive to modest changes in these parameters.
We used PyTorch and {\sc adam}\citep{paszke2017automatic,kingma2014adam}; the {\sc maml} code was adapted from \citet{pytorch-MAML}.  The code is available on \url{https://github.com/FerranAlet/modular-metalearning}.
The target output values $y$ in all data-sets were standardized to have mean 0 and standard deviation $1$.  The loss function then assigns loss 100 to a mean squared error of 1.  More experimental details are available in the supplement.

We tested these methods in three domains:  simple functional relationships, predicting the result of a robot manipulator pushing an object, and predicting the next frame of a kinematic skeleton based on previous frames using motion-capture data.  The last two domains represent the main motivation for this work:  a robot's experience of interacting with various entities in real-world situations should enable it to learn more efficiently about a new entity in a new situation.   There is typically some sensible decomposition of the prediction function, but that decomposition is not known in advance.  We hope that \algname{} can find an appropriate decomposition and that doing so will significantly leverage learning, as well as reveal interesting structure, in the new domain.

An additional advantage of \algname{} is computational efficiency.  Unlike {\sc maml}, it does not have extra gradient steps embedded in the inner loop at meta-training time, which offers some efficiency;  in addition, forward and backward passes can be done much more efficiently on GPUs by parallelizing over tasks.  {\sc maml}
is generally faster at online training time, since \algname{} has to search over structures. However, this training took at most 10 seconds in our examples. Moreover, to store a structure we only need a few integers, compared to storing a whole set of weights for parametric methods.

\subsection{Functions}
\label{sec:functions}
\begin{figure}[t]
    \centering
    {\includegraphics[width=0.345\textwidth]{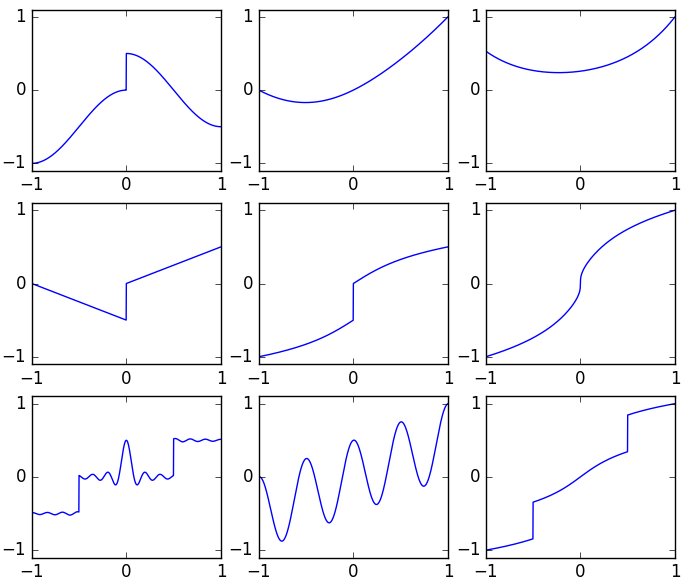}}\hfill 
    {\includegraphics[width=0.295\textwidth]{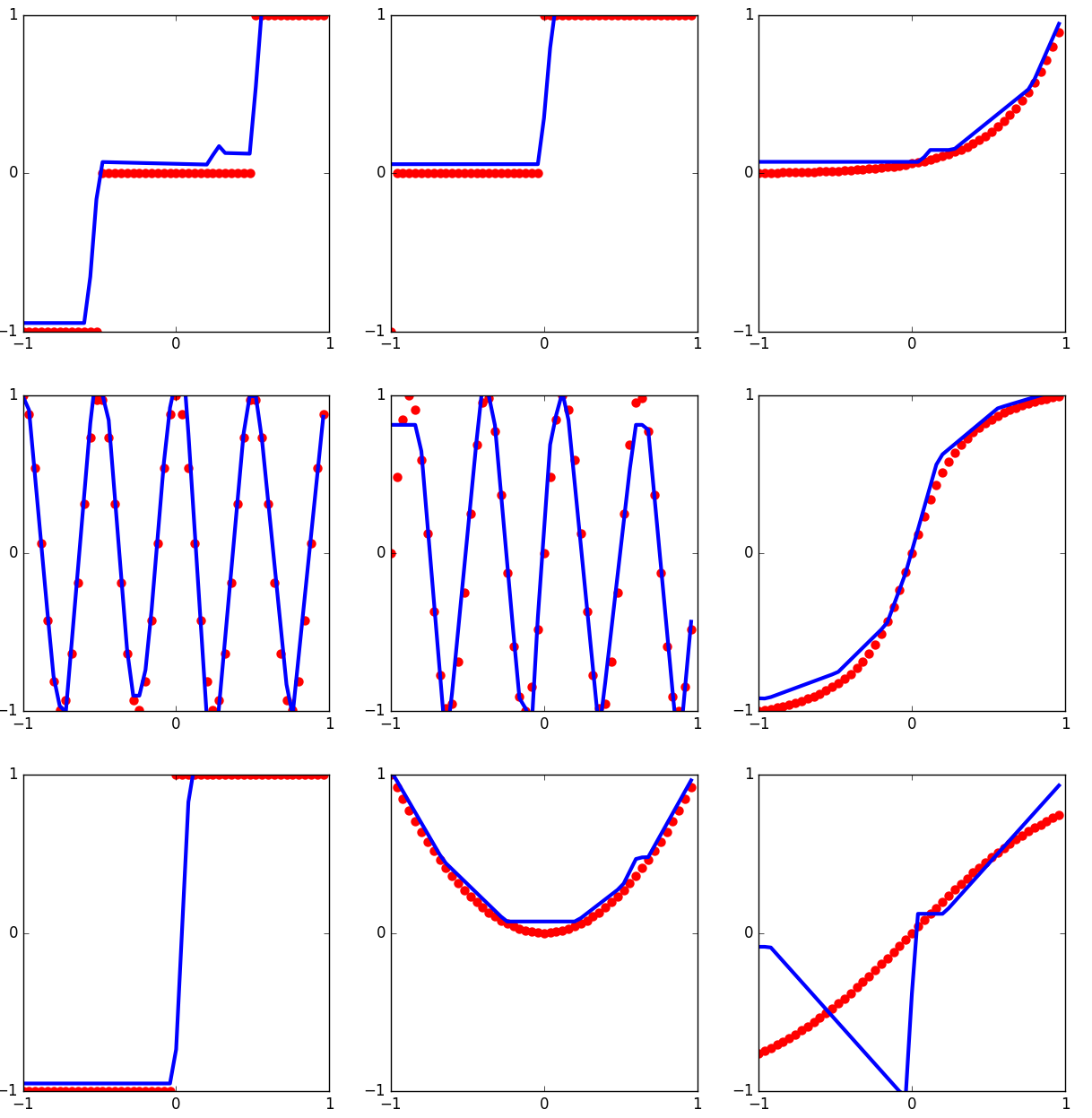}}\hfill 
    {\includegraphics[width=0.305\textwidth]{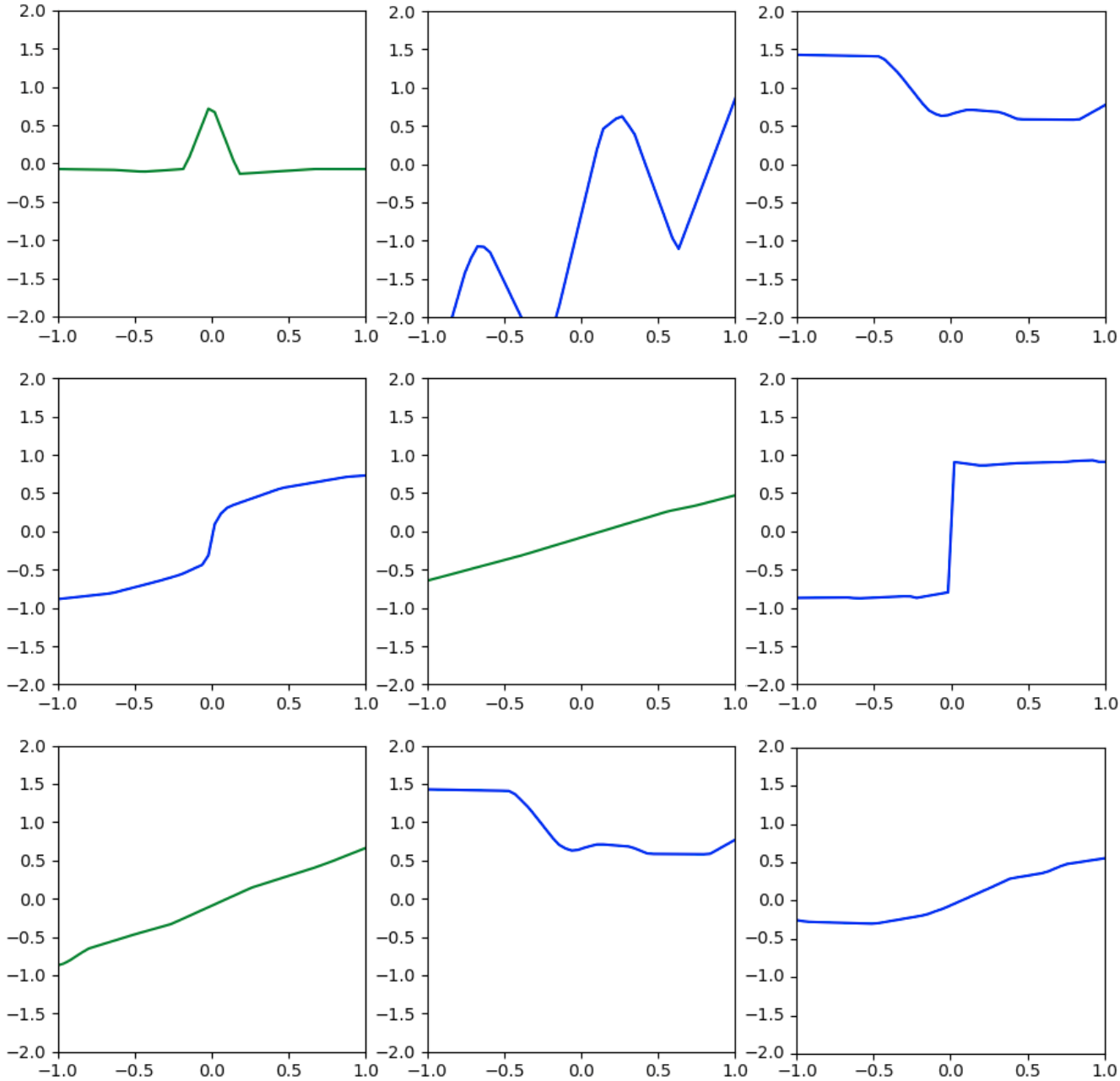}}
    \caption{Random functions (left); \algname{} (center) and \algMAML{} (right) modules. All but one \algname{} modules(blue) are nearly identical to a basis function(red).
\label{fig:basis_functions}}
\vspace{-1\baselineskip}
\end{figure}
We begin by testing on an extended version of the sine-function prediction problem~\citep{MAML}, which consisted of data-sets generated from functions of the form $\sin (ax + b)$ for varying values of $a$ and $b$.  
The compositional scheme for \algname{} is $h(x) = f_i(f_j(x))$;  $F$ consists of 20 feed-forward neural networks, 10 with 1 hidden layer and 10 with 2. In our experiments in this section {\sc maml} and {\sc Pooled} use the same architecture as the original {\sc maml} experiments.
We construct an additional illustrative domain consisting of sums of pairs of common non-linear functions, such as {\it exp} and {\it abs}, 
scaled so the output is contained in the range $[-1, +1]$, generating $16^2$ possible prediction tasks.  We use 230 randomly selected tasks for meta-training and a different task for testing. Example functions are shown on the left of figure~\ref{fig:basis_functions}.  
We use the same architectures for this domain as for the sine domain, except that the compositional scheme is $h(x) = f_i(x) + f_j(x)$.

The results are shown in table~\ref{table:results}.
As expected, training a single network on pooled data from all the tasks ({\sc Pooled}) works poorly in all of these domains.  In the sine domain, {\sc maml} outperforms \algname{} because the detailed parameter values are critical to performing well in a new domain, but \algMAML{} significantly improves on both methods, showing that both the structure and gradient meta-learning methods are useful.  For sums of functions, we report results in two cases: one in which we average over performance for 1--4 training examples, and one for 16 training examples.
With a small amount of online training data, \algname{} outperforms other methods because it has the proper structural prior. With more data, all methods improve, but \algname{} and \algMAML{} improve on {\sc maml}.
The plots in the middle and right of figure~\ref{fig:basis_functions} show some of the basis modules learned by \algname{} and \algMAML{}, respectively.  Those learned by \algname{} are an almost perfect recovery of the actual primitives used to generate the data, even though the algorithm had no information about those functions;  \algMAML{} has found similar functional shapes, yet with different values because it can still modify its parameters at online training time.

\begin{table}[t]
  \centering
  \vspace{8pt}
  \tabcolsep=0.11cm{
  \begin{tabular}{ccccccc}
    \toprule[1.5pt]
      \textbf{Dataset}& {\sc Pooled}& {\sc maml} & \textbf{\algname{}} & \textbf{\algMAML{}} & \textbf{Structure}\\
    \midrule[2pt]
      Parametrized sines & 98.1 &26.5 & 32.5 & \textbf{19.8} & composition\\ 
    \midrule
      Sum of functions (1-4pts)& 32.7 & 19.7 & \textbf{12.8} & 18.0 & sum\\
     \midrule
      Sum of functions (16pts) & 31.9 & 8.0 & \textbf{0.4} & \textbf{0.4} & sum\\
    \midrule[1.5pt]
      MIT push: known objects & 21.5 & 18.5 & 16.7 & \textbf{14.9} & attention\\
    \midrule
      MIT push: new objects  & 18.4  &\textbf{ 16.9 }& \textbf{17.1} & \textbf{17.0} & attention\\
    \midrule[1.5pt]
      Berkeley MoCap: known actions  & 35.7 & 35.5 & \textbf{32.2} & \textbf{31.9} & concatenate\\
    \midrule
      Berkeley MoCap: new actions & 79.5 & 77.7 & 77.0 & \textbf{73.8} & concatenate\\ 
    \bottomrule[1.5pt]
  \end{tabular}}
  \caption{Summary of results; lower is better; bold results are not significantly different from best.}
  \label{table:results}
  \vspace{-2\baselineskip}
\end{table}

\subsection{Learning to model results of pushing actions}

An important sub-problem in robot manipulation is to acquire models of the effects of the robot's motor actions on the objects in the world.  
The MIT push data-set \citep{yu2016more} contains the results of executing pushing actions with a manipulator hand, for 11 different objects with different shapes on 4 surfaces with different friction properties.  The behavior of the object on these surfaces is close to quasi-static 
, so the state can be characterized by an input $x$ consisting of: position of the object (2d), orientation of the object (1d), position of the pusher (2d), and velocity of the pusher (2d).  Given this 7-d input, the objective is to predict the 3-d change in the object's position and orientation.   Each task represents experience with a particular object on a particular surface. 

The compositional scheme for \algname{} is the weighted ensemble described in section~\ref{sec:compositions};  $F$ consists of 20 feedforward neural networks, 10 attention modules and 10 regressors.  
We consider two different meta-learning scenarios.  In the first, the object-surface combination in the test case was present in some meta-training task;  in the second, the objects used in the meta-training tasks differ from the object used in the test task.  The results in table~\ref{table:results} show that, for previously encountered objects, \algMAML{} performs best, and \algname{} outperforms {\sc maml}.  For unknown objects, all three approaches perform roughly equivalently.  

\begin{figure}[t]
    \centering
    {\includegraphics[width=0.47\textwidth]{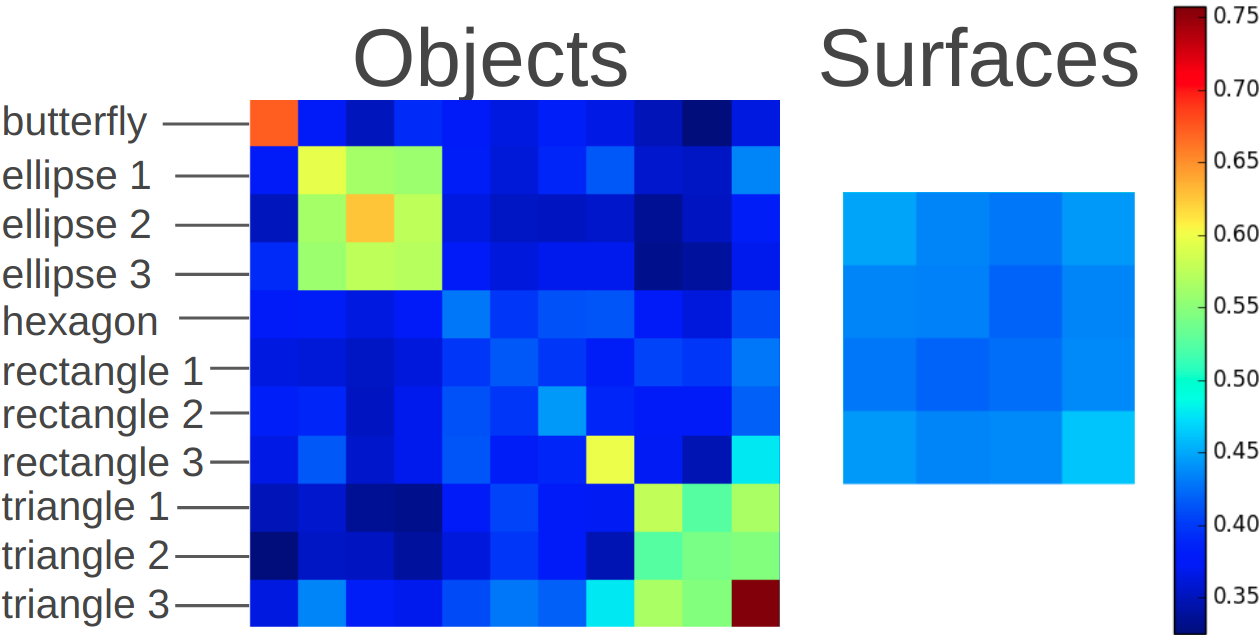}}\hfill
    {\includegraphics[width=0.51\textwidth]{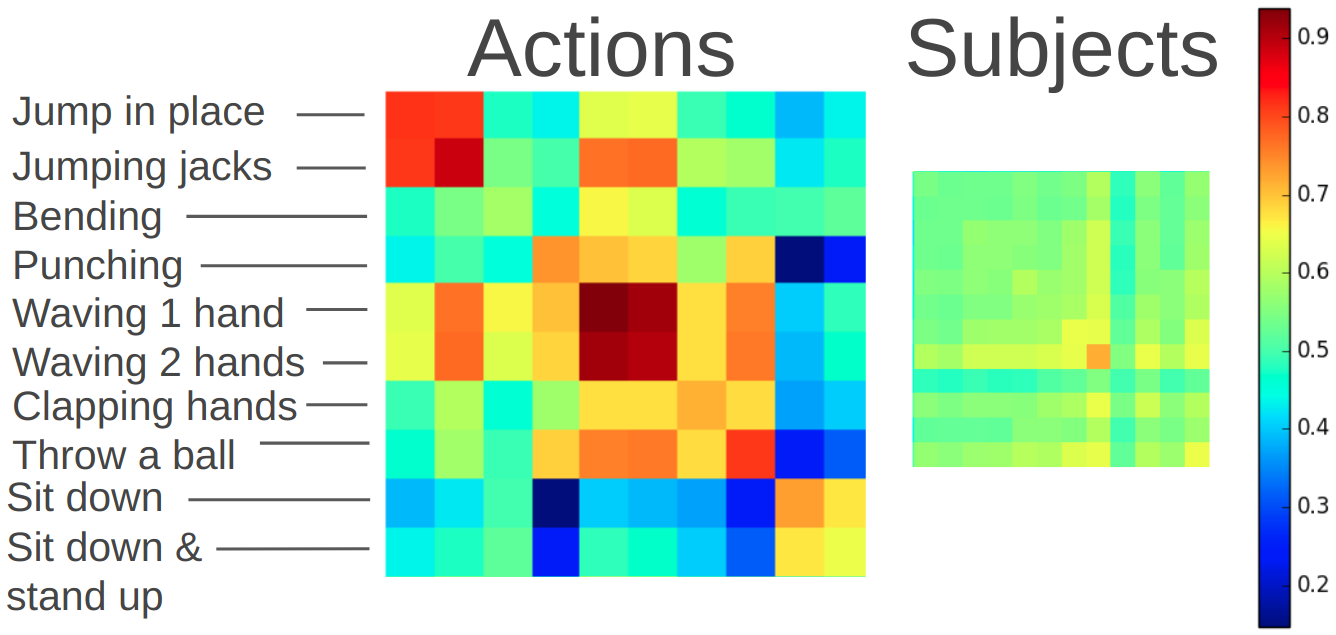}}

    \caption{Shared modules show internal structure of the datasets (left--pushing; right--motion). In particular, less important factors (surfaces and subjects) do not change the structure while bigger changes (objects and actions) do. Within the structural changes, there is more sharing between conceptually similar datasets. \label{fig:shared_modules}}
    \vspace{-\baselineskip}
\end{figure}

Another important aspect of the structured hypothesis space is that it can give us insight into the relationships between tasks.  Figure~\ref{fig:shared_modules} illustrates the structural relationships that were uncovered in this data.  The matrix on the left is indexed according to which object was being pushed.  The entry in location $i, j$ represents the average percentage of modules shared by the structure learned to predict results for object $i$ and the structure learned for object $j$.  We can see it distinguishes 3 clusters of data: butterfly, all ellipses, and all triangles. The biggest rectangle shares modules with the biggest triangle, probably due to similar size and aspect ratio.  The matrix labeled "Surfaces" does not show dependence on the surface, as expected for the quasi-static regime.

\subsection{Predicting skeleton configurations}

Robots that interact with humans should be able to understand and predict human action, both for the purposes of safety and for task-driven collaboration.  We applied meta-learning to this problem, using data from the Berkeley MHAD motion capture dataset~\citep{ofli2013berkeley}.  This domain is dynamic, and so we use three previous configurations (at intervals of 0.1 sec) of a human skeleton to predict the next one.  Each configuration is characterized by one 3-d position and 90 joint angles describing a kinematic tree, so the input has dimension 279 and the output has dimension 93.  
There are 12 subjects performing 11 different actions several times, for a few seconds each. 

We constructed a compositional scheme for \algname{} that is related to this task.  It has a fixed first layer with 128 output units to compress the input, which is the same for all structures, and independent ``parallel'' modules that take those 128 inputs and produce kinematic sub-trees for each body part (2 legs, 2 arms, and torso).  
For {\sc maml} and {\sc Pooled} we use a single architecture of the same depth with 4 times more parameters.
We again consider two different meta-learning scenarios.  In the first, the activities used in the training task are the same as the activities used in the meta-test task, but the human actor varies;  in the second, the activities used in the training tasks differ from the activity used in the test task.  The results in table~\ref{table:results} show that, for known activities, \algname{} and \algMAML{} perform best.  For unknown activities, none of the methods perform very well, but \algMAML{} outperforms the others. 
We obtain a similar pattern of correlation among shared modules, shown in figure~\ref{fig:shared_modules}, in which there is significant module-sharing among similar tasks and no real pattern of module-sharing among human actors.


\subsection{Conclusion} We have demonstrated that modular compositional structure can provide a useful basis for transferring learned competence from previous tasks to new tasks. It can also yield insight into the underlying structure of the domain.  We believe this \textit{combinatorial generalization} is a promising route to scale to large numbers of tasks and continual learning settings as we can increase our knowledge in modular ways without forgetting previously learned concepts. The structural information to be provided in advance is a few lines of code to describe the possible modifications that can be made to a structure, which is not much more than would be required for specifying a typical neural network.
\clearpage
\acknowledgments{We gratefully acknowledge support from NSF grants 1420316, 1523767 and 1723381 and from AFOSR grant FA9550-17-1-0165. F. Alet is supported by a La Caixa fellowship. Any opinions, findings, and conclusions or recommendations expressed in this material are those of the authors and do not necessarily reflect the views of our sponsors. \\We want to thank Zi Wang for her assistance in setting up the experiments, Maria Bauza for her help with the MIT push dataset and Rohan Chitnis for his comments on an initial draft. Finally, we thank reviewers for their useful suggestions.}


\bibliography{references}  
\newpage
\appendix

\section{More insight into the difficulty of meta-training}
Once the modules are trained, finding the best structure is just a matter of search. Similarly, if someone told us the best structure for each task, we would be able to find the best parameters by pure gradient descent. However, we start in the opposite situation: we don't know the module weights nor the best structure for each dataset. This leads to a chicken-and-egg problem: the concept of best structure is meaningless without first having good modules and we cannot train the modules if we do not know which roles they should play.

\begin{figure}[h!]
    \centering
    \includegraphics[width=\textwidth]{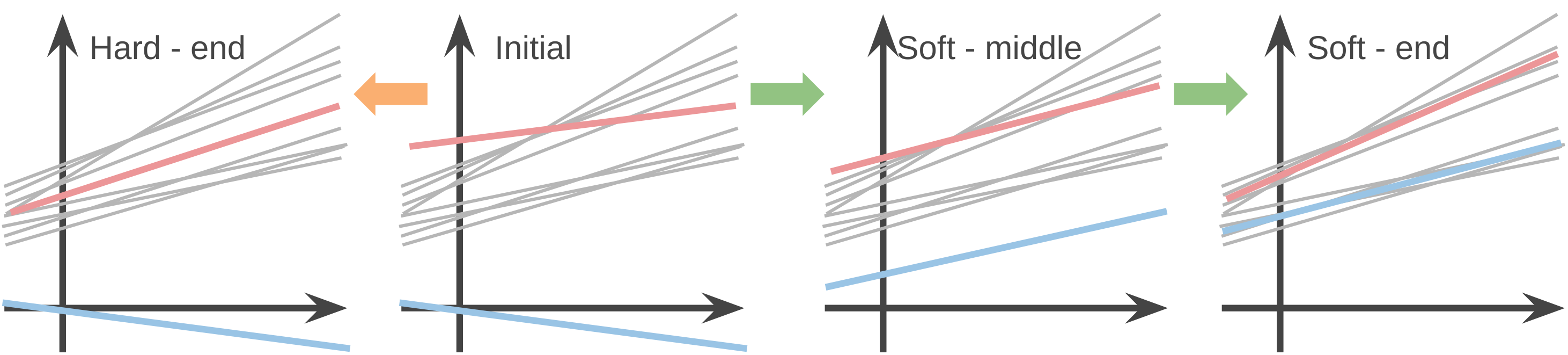}
    \caption{Gray lines are different tasks; the composition is just a single module $h(x)=f_i(x)$, either the red or the blue. We have to place our modules such that they cover the tasks (grey lines well). The second frame represents the initial state of a search for parameters.  If we make local steps in structure and parameter space, we will converge to the solution on the left, without ever updating  the blue.  However, if we consider a smoothed criterion with a non-point distribution over structures, we will update the parameters for the blue module and eventually arrive at the solution on the far right. \label{fig:comparative_advantage}} 
    \vspace{-1.5\baselineskip}
\end{figure}

An important problem this causes, illustrated in figure \ref{fig:comparative_advantage} is that if we greedily optimize the structure we have the risk of premature optimization and running into a local optima. This motivated our smoothed objective where modules and structures slowly adapt to one another.
\section{More results on functions dataset}
In the main text we claim we find the basis set of functions. This is compatible with some modules not having a closeby function, since there are 20 modules for 16 basis functions. To prove our claim, we plot the 16 functions and the closest module to each of them.
\begin{figure}[h!]
    \centering
    \includegraphics[width=\textwidth]{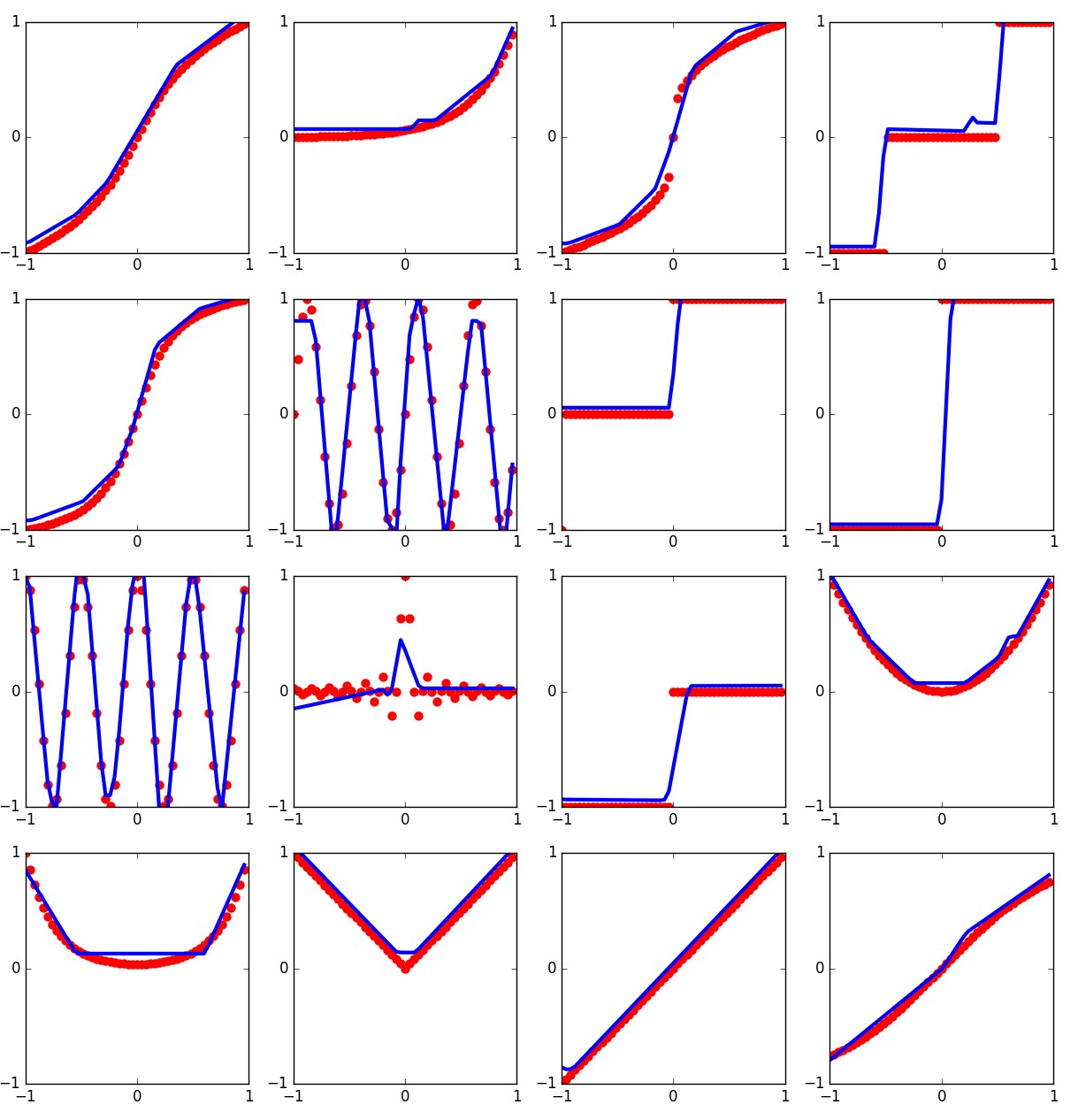}
    \caption{Our learned set of modules recovers all 16 basis functions. The 16 basis functions (never seen alone by the algorithm) are plotted in red, the closest module is plotted in blue. All except the since function near 0 are close matches. All modules are different except for row 2, column 3 and 4: {\tt $\sqrt[3]{x}$} and {\tt $arctan(4*x)/\pi$} }
    \label{fig:closest-module}
\end{figure}
\begin{figure}[h!]
    \centering
    \includegraphics[width=.5\textwidth]{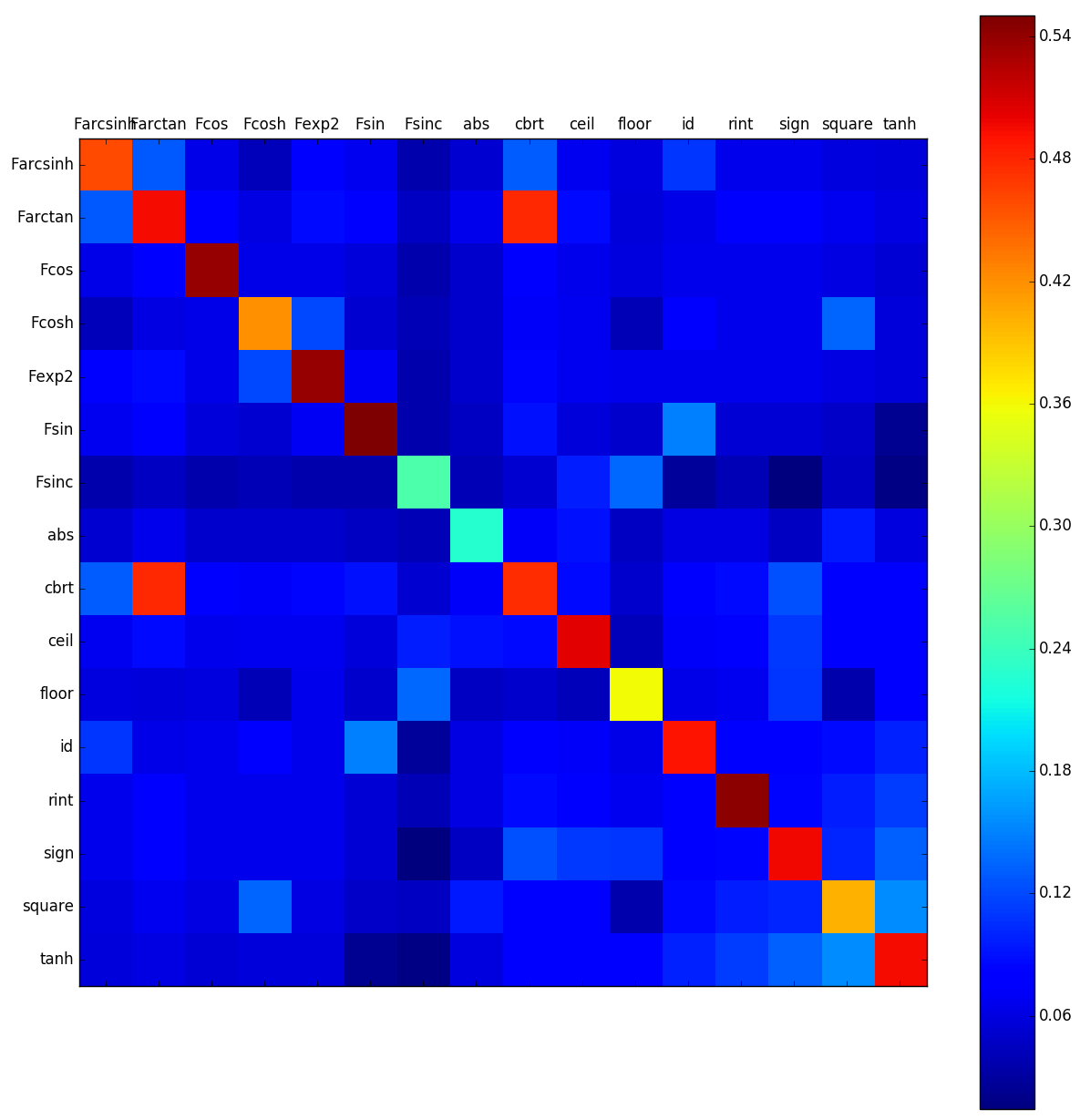}
    \caption{Sharing between datasets containing each function, similar to figure \ref{fig:shared_modules}. The diagonal is very dominant, showing if two datasets which share one function their corresponding structures will likely share a module. Only $sinc(x)$ and $|x|$ don't have near the predictable $50\%$ sharing rate: one because the fitted module is not perfect, the other because it has two modules that fit it perfectly. The other exception is the relation between $\arctan{x}$ and $\sqrt[3]{x}$, since a single module is close to both of them.}
    \label{fig:sharing-rate-fns}
\end{figure}
\begin{figure}[h!]
    \centering
    \includegraphics[width=.5\textwidth]{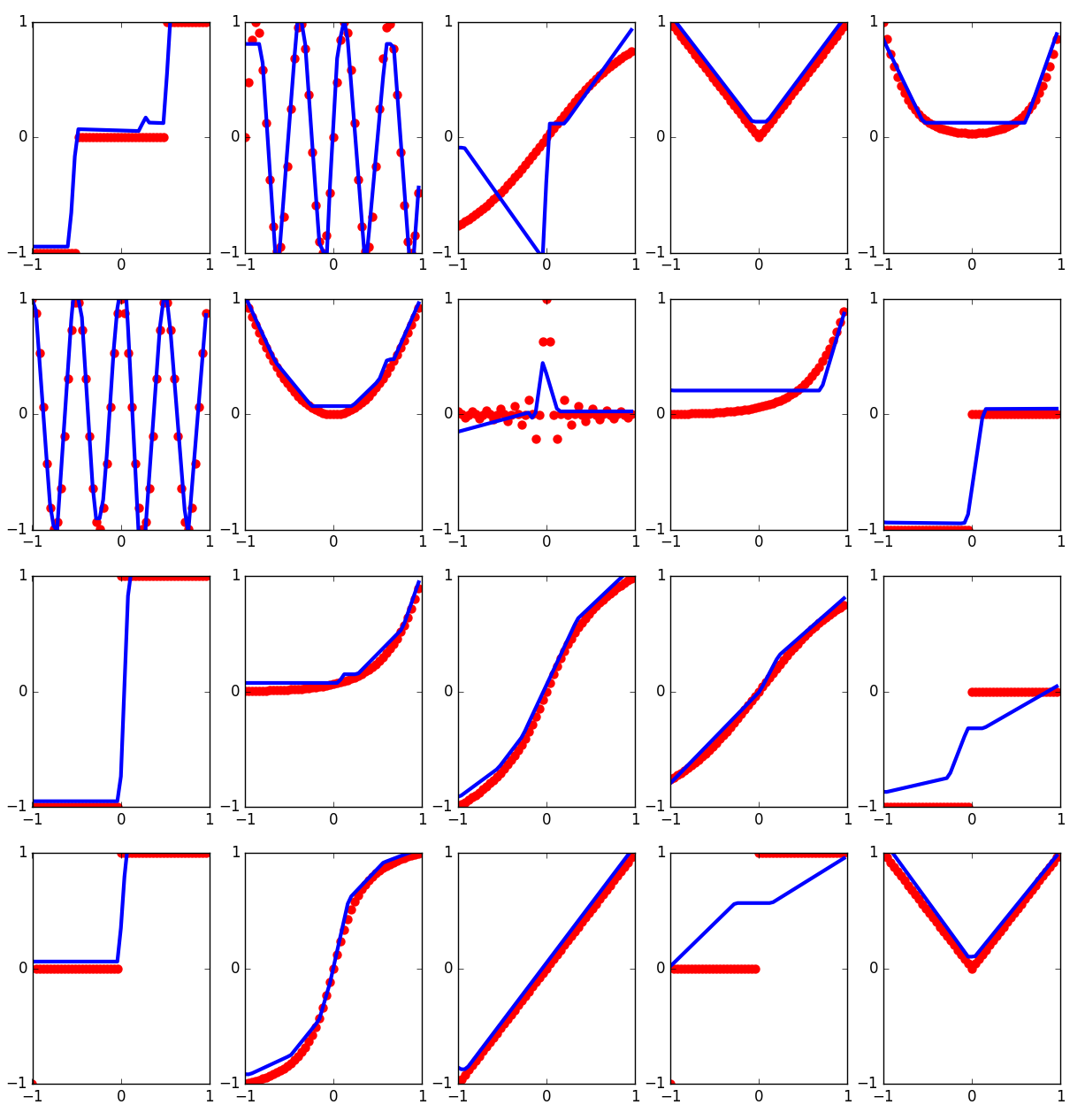}
    \caption{All modules and their closest function, completing figure \ref{fig:basis_functions}. There are 20 modules: 4 are useless, 2 encode $|x|$, 13 encode a single function, 1 encodes both the $\sqrt[3]{x}$ and $arctan(4*x)/\pi$.}
    \label{fig:my_label}
\end{figure}
\section{Experimental details}
Functions used in the functions dataset: {\tt ($abs$, $arcsinh(4x)/arcsinh(4)$, $arctan(4x)/arctan(4)$, $cbrt$, $ceil$, $cos(4\pi x)$, $cosh(4x)/cosh(4)$, $exp2(4x)/exp2(4)$, $floor$, $rint$, $sign$, $sin(4\pi x)$, $sinc(4\pi x)$, $square$, $tanh$, $id$)}. To create a dataset we picked all pairs of functions and More information, including the dataset itself, can be found in \url{http://lis.csail.mit.edu/alet/modular-metalearning.html} and \url{https://github.com/FerranAlet/modular-metalearning}.

\begin{table}[h!]
  \centering
  \vspace{8pt}
  \tabcolsep=0.11cm{
  \begin{tabular}{c|c|c|c|c}
    \toprule[1.5pt]
      \textbf{Dataset}& sine & functions & MIT push & Berkeley MoCap \\
    \midrule[2pt]
      \# training metatasks &230 & 230 & 236 & 236\\
      \# training points & 16 & (1-4),16 & 32 & 128\\
      \# validation points & 64& 64 & 48 & 192\\
    \midrule[1.5pt]
      \#nodes {\sc Pooled} \& {\sc maml} & [1-64-64-1] &  & [7-128-64-3] & [279-512-97]\\
      \#modules per type of module & 10,10&  & 10,10&1,3,3x4\\
      \#nodes \OUR \& \algMAML & [1-16-1], & same as left & [7-32-1], &[279-128],[128-21],\\
      & [1-16-16-1] & & [7-64-32-1] & [128-18]x4\\
      learning rate for all architectures & 0.003 & & 0.001 & 0.001 \\
     \midrule[1.5pt]
     statistical variability & 0.8 & 0.4 & 0.7 & 0.7 \\
    \bottomrule[1.5pt]
  \end{tabular}}
  \caption{Summary of number of training plus architectural descriptions. \label{table:architectures}}
\end{table}
Learning rates and epochs were generally the same. {\sc Pooled} and \OUR\ had twice as many epochs in MIT push and Berkeley (500 vs 1000), still taking less amount of time to train thanks to being  4 times faster. We tried several similar architectures and learning rates for all algorithms and checked all algorithms converged appropriately. Other parameters: MAML inner updates: 5, MAML step size 0.001. For an up-to-date version of the implementation please visit \url{https://github.com/FerranAlet/modular-metalearning}.
\end{document}

%% file: variables.tex
\usepackage{graphicx}
\usepackage[]{todonotes}

\usepackage{algorithm}
\usepackage[noend]{algpseudocode}
\usepackage{tabularx}
\usepackage{comment}

\usepackage{subfig}
\usepackage{caption}
\usepackage{booktabs}

 \newcommand{\add}[2][]{}